\documentclass{article} %
\usepackage{colm2024_conference}

\usepackage{booktabs}
\usepackage{graphicx}
\usepackage{enumitem}
\usepackage{wrapfig}
\usepackage{algorithm}
\usepackage{algpseudocode}

\usepackage{microtype}
\usepackage{amsmath}
\usepackage{amsfonts}
\usepackage{colortbl}
\usepackage[utf8]{inputenc}
\usepackage[T1]{fontenc}
\definecolor{lightgray}{rgb}{0.9,0.9,0.9}
\usepackage{caption}
\usepackage{subcaption}
\usepackage{xcolor}
\usepackage{setspace}
\usepackage{url}
\usepackage{multirow}
\usepackage{colortbl}
\usepackage{tabularx}
\usepackage{blindtext}
\usepackage{pgfplots}
\pgfplotsset{compat=1.18}
\usepackage{tikz}
\usetikzlibrary{er,positioning,bayesnet}
\usepackage{makecell}
\usepackage{tipa}
\usepackage{siunitx}
\usepackage{nicefrac}
\usepackage{tocloft}
\usepackage{listings}
\usepackage[raster,skins]{tcolorbox} %
\usepackage{xltabular}
\usepackage{adjustbox}
\usepackage{xurl}
\usepackage{xcolor}

\usepackage{amsmath,amsfonts,bm}

\def\eqref#1{equation~\ref{#1}}

\def\1{\bm{1}}

\DeclareMathAlphabet{\mathsfit}{\encodingdefault}{\sfdefault}{m}{sl}
\SetMathAlphabet{\mathsfit}{bold}{\encodingdefault}{\sfdefault}{bx}{n}

\title{Unlocking Cognitive Capabilities and Analyzing the \\Perception-Logic Trade-off}

\author{
  \vspace{4pt}
  \textbf{MERaLiON Team} \\
  \vspace{0.5em}
  \begin{tabular}{c}
    Corresponding Author: Longyin Zhang \\
    \noalign{\vspace{0.2em}}
    Institute for Infocomm Research (I$^2$R), A*STAR, Singapore \\
    \noalign{\vspace{0.2em}}
    \texttt{Zhang\_Longyin@a-star.edu.sg}
  \end{tabular}
}

\begin{document}

\maketitle

\begin{abstract}
  Recent advancements in Multimodal Large Language Models (MLLMs) pursue omni-perception capabilities, yet integrating robust sensory grounding with complex reasoning remains a challenge, particularly for underrepresented regions. In this report, we introduce the research preview of \textbf{MERaLiON2-Omni (Alpha)}, a 10B-parameter multilingual omni-perception tailored for Southeast Asia (SEA). We present a progressive training pipeline that explicitly decouples and then integrates ``System 1'' (Perception) and ``System 2'' (Reasoning) capabilities. First, we establish a robust Perception Backbone by aligning region-specific audio-visual cues (e.g., Singlish code-switching, local cultural landmarks) with a multilingual LLM through orthogonal modality adaptation. Second, to inject cognitive capabilities without large-scale supervision, we propose a cost-effective \textbf{Generate-Judge-Refine} pipeline. By utilizing a Super-LLM to filter hallucinations and resolve conflicts via a consensus mechanism, we synthesize high-quality silver data that transfers textual Chain-of-Thought reasoning to multimodal scenarios.

  Comprehensive evaluation on our newly introduced \textbf{SEA-Omni Benchmark Suite} reveals an \textbf{Efficiency-Stability Paradox}: while reasoning acts as a non-linear amplifier for abstract tasks (boosting mathematical and instruction-following performance significantly), it introduces instability in low-level sensory processing. Specifically, we identify \textbf{Temporal Drift} in long-context audio, where extended reasoning desynchronizes the model from acoustic timestamps, and \textbf{Visual Over-interpretation}, where logic overrides pixel-level reality. This report details the architecture, the data-efficient training recipe, and a diagnostic analysis of the trade-offs between robust perception and structured reasoning.
\end{abstract}

\begin{figure}[tbh]
    \centering
    \includegraphics[scale=0.5]{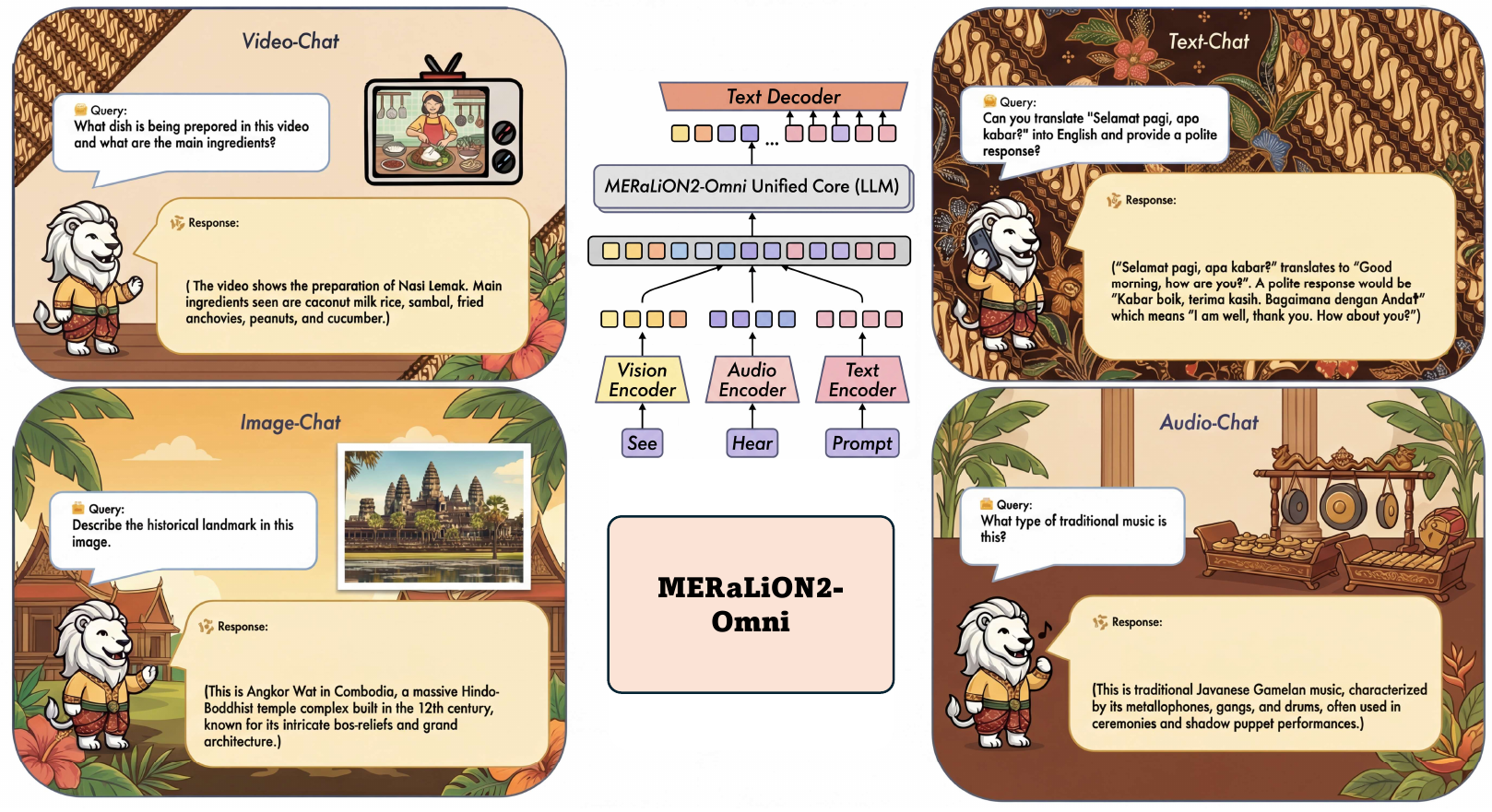}
    \caption{MERaLiON2-Omni (Alpha) is a 10B omni model for Southeast Asia that extends MERaLiON-v2 with a stronger perception backbone aligning region-specific audio-visual signals to a multilingual LLM, and a Cognitive Reasoning Layer enabled by Chain-of-Thought transfer.}
    \label{fig:ovewview_arch}
\end{figure}

\section{Introduction}
\label{sec:intro}

The pursuit of ``Omni'' models, unified systems capable of perceiving and reasoning across text, audio, and vision, has largely been driven by scaling laws~\citep{xu2025qwen2, xu2025qwen3, bai2023qwen, caffagni2024revolution, yin2024survey}. Leading proprietary models demonstrate that massive data scale and parameter counts can eventually bridge the modality gap.
However, for research teams operating on limited compute budgets, particularly in underrepresented regions like Southeast Asia (SEA), relying solely on scale is neither computationally feasible nor culturally effective~\citep{nguyen2024seallms, Zhang2025SeaLLMs3, tjhi-etal-2023-sea, pouget2024no, kadiyala2025uncovering}. Under these constraints, we hypothesize that the central challenge is not scaling further, but to find efficient ways to carry over LLM reasoning into multimodal settings without restarting pre-training.

In this report, we introduce \textbf{MERaLiON2-Omni} (Alpha), a family of 10B omni models designed for the SEA context. In particular, we design the training pipeline as two progressive conceptual stages to achieve data efficiency. First, we build a \textbf{Multilingual Perception Backbone (System 1)} that aligns region-specific audio signals (e.g., Singlish code-switching)~\citep{wang2025advancing, he2025meralion} and local visual cues (such as food or landmarks)~\citep{cahyawijaya-etal-2025-crowdsource} with a foundational language model. On top of this foundation for cross-modal perception, we add a \textbf{Cognitive Reasoning Layer (System 2)} to encourage our omni model to move beyond passive perception and engage in explicit multi-step reasoning and logical composition.

Following this perception–reasoning duality, we first introduce the foundational \textbf{Perception Backbone}. In brief, the backbone uses a modular encoding design that combines a SigLIP vision encoder~\citep{bai2023qwen} and the Whisper-based audio encoder from our previous work of MERaLiON-v2~\citep{he2025meralion}.
Our findings reveal a near-perfect \textbf{orthogonality} between audio and visual perception capabilities (parameter updates), motivating our plan of aligning visual modalities to our pre-trained Audio-LLM without severely damaging the original audio-text knowledge~\citep{liang2022mind}. On this basis, we obtain the ``System 1'' with stable sensory grounding needed to support our exploration of multi-modality reasoning.

Given limited access to large-scale multimodal reasoning data, we introduce a \textbf{two-stage reasoning injection} approach to achieve ``System 2''. Firstly, we train the omni model with massive textual reasoning data to encourage creative reasoning behaviors, though this sometimes leads to hallucinations due to the notorious reasoning drift problem ~\citep{shumailov2024ai}. Secondly, we further utilize a silver data refinement pipeline with a strong LLM judge (Qwen-3-235B)~\citep{yang2025qwen3} prompted to compare multiple reasoning paths and favor facts supported by consensus. This refined data allows us to transfer textual reasoning into multimodal settings more safely without the need for large-scale pretraining.

We conduct a series of quantitative and qualitative experimental analyses in this report to explore the interesting phenomenon of \textbf{Efficiency–Stability Paradox} where reasoning acts as a non-linear amplifier of modality understanding. More specifically, it enhances performance on abstract tasks (e.g., +11.2\% on Math) by mapping richer visual and audio inputs into the language space, but it also introduces instability in low-level perception. Finally, we identify two failure modes as follows:
\begin{itemize}
  \item \textbf{Temporal Drift:} We find that reasoning helps interpretation in short audio clips (<10s), but under long-context stress tests (>30s), extended Chain-of-Thought sequences gradually desynchronize from acoustic timestamps.
  \item \textbf{Visual Over-interpretation:} In static visual tasks, the pressure to maintain logical consistency sometimes leads the model to over-interpret scenes, constructing plausible narratives that are weakly supported by the actual pixels.
\end{itemize}

In light of these observations, we release the current version as a \textbf{research alpha}. We emphasize that this work serves primarily to dissect the intrinsic trade-offs between \textbf{robust sensory grounding} and \textbf{complex reasoning}, rather than to present a fully polished industrial solution.

The remainder of this report is organized as follows: Section 2 details the modular architecture. Section 3 introduces our progressive training methodology, shifting from perception alignment to reasoning injection. In Section 4, we describe the data construction and the SEA-Omni Benchmark. Finally, Section 5 presents extensive experimental results and analyses, followed by related work and conclusions.

\section{Architecture}
Adhering to a modular design philosophy, we prioritize parameter efficiency and flexibility when modeling our unified, end-to-end MERaLiON2-Omni system. This allows for the independent optimization of our used encoders before unified alignment. The final architecture consists of three primary components, the SEA-aligned language backbone, the audio encoder, and the visual encoder, connected via trainable projectors.

\subsection{Multilingual Language Backbone}
Following our previous work~\citep{he2025meralion}, we continue to use the decoder-only \textbf{Gemma-2-9B} model~\citep{team2024gemma} as our central reasoning unit. Considering the linguistic diversity of Southeast Asia, we utilize a variant of the model that has undergone continuous pre-training on a region-specific corpus covering multiple SEA languages~\citep{he2025meralion}. In this way, it ensures that the unified omni architecture possesses SEA priors before any multimodal tokens are introduced.

\subsection{Audio Perception Module}
Consistent with our prior work on SEA Audio-LLM~\citep{he2025meralion}, we integrate a \textbf{Whisper-based encoder (Large-v3)}~\citep{radford2023robust} for auditory perception and project the continuous audio representations directly into the LLM's embedding space via a \textbf{Convolutional Adapter (Conv1d)}. Considering the balance between training resources and audio processing quality, we set the audio encoder to process audio at a resolution of \textasciitilde3.3 Hz after downsampling (i.e., one token per 300ms). It allows the model to capture paralinguistic features like tone, emotion, and background noise alongside linguistic content, providing a stable basis for the final omni model.

\subsection{Visual Perception Module}
Visual inputs are processed using a Vision Transformer (ViT) initialized from the open-source \textbf{Qwen2.5-Omni checkpoint}~\citep{xu2025qwen2}. To achieve data efficiency, rather than training the vision encoder from scratch, we directly inherit the weights of a mature visual encoder and project image patches through a ``Patch Merger'' (MLP) to align their dimensions with the LLM. This proposed visual pipeline can handle both static images and video frames where videos are treated as temporal sequences of images, allowing the model to perform spatial reasoning and temporal visual analysis within a unified visual embedding space~\citep{xu2025qwen2, bai2023qwen}.

\vspace{0.2cm}
\noindent
In our final omni model, audio segments and visual patches are converted into discrete ``pseudo-tokens'' that occupy the same embedding space as language tokens. Moreover, these tokens are interleaved sequentially in the context window. For instance, a video input is finally represented as a sequence of tokens comprised of \texttt{<VideoHere>}, \texttt{<AudioHere>}, and \texttt{<Text Instruction>}. Through this unified representation, our omni model can perform standard autoregressive generation conditioned on any arbitrary sequence of text, sound, and sight, without requiring complex specialized attention mechanisms.

\section{Training Methodology}
We design the training pipeline in a progressive fashion, with the most basic intuition that robust perception must be established before complex reasoning.
Under this philosophy, we evolve the model from an omni-perception baseline into a reasoning-enhanced version. Formally, we maintain a standard objective to maximize the likelihood of next-token generation given a multimodal context $X = \{X_v, X_a, X_t\}$, where $v$, $a$, and $t$ refer to visual, audio, and textual modalities, respectively.

\subsection{Phase 1: Establishing the Perception Backbone}
We adopt a three-stage progressive training for the basic MERaLiON2-Omni model, which focuses on optimizing the geometric structure of cross-modal alignment and data efficiency.

\textbf{Efficient Vision-Language Alignment.} We first build a semantic bridge between the vision encoder and our previously trained Audio-LLM~\citep{he2025meralion}. To achieve it, we introduce Low-Rank Adaptation (LoRA)~\citep{hu2022lora} matrices $\Delta\Theta_V$ and $\Delta\Theta_L$ for the vision encoder (including the projector) and LLM, respectively. On this basis, we train them on general-domain image-text pairs using the following objective:

\begin{equation}
\label{eq:stage2}
 \min_{\Delta\Theta_V, \Delta\Theta_L} \mathbb{E}_{(v,t) \sim \mathcal{D}_{g\_img}} [\mathcal{L}_{CE}(t|v; \Theta_V + \Delta\Theta_V, \Theta_L + \Delta\Theta_L)]
\end{equation}

Our heatmap suggests that improving visual performance operates in a direction largely independent of audio capability. This geometric finding allows us to align visual modalities to the foundation model without catastrophic interference with its pre-existing auditory capabilities~\citep{liang2022mind}.

\textbf{Deep Unimodal Visual Knowledge Injection.} We argue that a key innovation for regional adaptation should be the targeted refinement of the vision encoder. In this work, we merge the LoRA weights from stage 1, then \textbf{freeze all parameters except the vision encoder $\Theta_V$}. Afterwards, we perform full-parameter fine-tuning on our curated \textbf{SEA-Image} dataset. This encourages our vision encoder to become a domain expert on region-specific visual concepts, which can effectively distinguish subtle differences, for example \textit{Nasi Lemak} vs. \textit{Nasi Kandar}, without severely disturbing the LLM's original language distribution.

\textbf{Advanced Multimodal \& Multilingual Fusion.} Finally, we integrate all the three modalities (Audio, Image, Video, and Text) into a unified multilingual alignment. Previous iterations were largely limited to English instructions. We extend this stage by constructing a \textbf{multilingual multimodal instruction-following} dataset. To achieve this, we simply utilize an automated translation pipeline to expand original English instruction-tuning data into six additional regional languages, i.e., Mandarin, Indonesian, Malay, Thai, Vietnamese, and Tamil.

To achieve the unified alignment, we activate LoRA parameters for all components ($\Delta\Theta_V$, $\Delta\Theta_A$, and $\Delta\Theta_L$) and train them on this mixed multilingual instruction-following corpus as below.

\begin{equation}
\min_{\Delta\Theta} \mathbb{E}_{(v, a, t_{lang}) \sim \mathcal{D}_{multi}} [\mathcal{L}_{CE}(t_{lang}|v, a; \Theta + \Delta\Theta)]
\end{equation}

To this end, we establish our ``System 1'' perception backbone, which will be further utilized as a frozen foundation to ground our subsequent reasoning experiments.

\subsection{Phase 2: The Reasoning Shift}
To pivot from the ``System 1'' perception engine to the cognitive ``System 2'' omni model, we implement a cost-effective, two-step Reasoning Injection pipeline: (1) we first push the model to unlock its creative reasoning potential via pure reasoning texts, and (2) then constrain it with sensory evidence from our multimodal synthetic workflow.

\subsubsection{Cognitive Warm-up}
We introduce a step called ``\textbf{Creative Warm-up}'' to encourage the model to produce longer reasoning chains.
To improve the model’s ability to handle complex logical flows, we curate a subset of textual Chain-of-Thought (CoT) data from the AM-DeepSeek-R1-Distilled collection~\citep{zhao202514millionopensourcedistilled} to post-train the obtained ``System 1''. To prevent early training instability, we adopt a Curriculum Learning strategy~\citep{bengio2009curriculum} where we explicitly prioritize shorter reasoning chains (L < 512 tokens) before scaling up.

\begin{equation}
\min_{\Delta\Theta_{\text{reason}}} \mathcal{L}_{\text{CoT}}(t_{\text{reason}}|t_{\text{prompt}})
\end{equation}

This step improves textual reasoning. However, it can also introduce uncontrolled creativity.
When directly applied to multimodal inputs, this model tends to hallucinate, sometimes inventing visual details to satisfy its internal logic~\citep{shumailov2024ai, chen2025multimodal}, which we address in the following step.

\subsubsection{Silver Data Alignment}
We synthesize a \textbf{Silver Multimodal Reasoning Dataset} to anchor model outputs in actual sensory evidence.
In practice, we use the above model as a generator and rely on the pre-trained Qwen-3-235B Judge~\citep{yang2025qwen3} to filter and refine the outputs. The Generate-Judge-Refine workflow works as follows:

\textbf{1. Divergent Generation:} We prompt the initial reasoning model to generate $K$ diverse reasoning paths $C = \{c_1, ... c_K\}$ for a given image, video, or audio input $X$.

\textbf{2. Convergent Refinement (Consensus \& Blurring):} The LLM Judge $\mathcal{M}_{\text{judge}}$ evaluates the candidates using a \textbf{Blur-and-Confirm} method:
\begin{itemize}
    \item \textbf{Confirmation}: When the majority of candidates agree on a specific sensory detail (e.g., ``the car is red''), the Judge confirms this as a high-confidence fact.
    \item \textbf{Conflict Blurring}: When candidates hallucinate conflicting details (e.g., one says ``blue hat,'' another says ``green hat''), the Judge is instructed to ``blur'' the reasoning into a generic statement (e.g., ``a person wearing a hat'') or explicitly state uncertainty.
\end{itemize}

Using this workflow, we obtain a high-quality target sequence $Y_{\text{silver}}$ that maintains logical structure with limited perceptual hallucination.

\textbf{3. Stopping Signal Injection:} To further prevent the ``endless thinking'' loops observed in many reasoning models, we inject explicit \textbf{Reasoning Delimiters} (i.e., `<think>...</think>`) into the silver data. This teaches the model exactly when to switch from internal deliberation to external response generation.

Lastly, we perform lightweight LoRA fine-tuning on this carefully curated training data to yield \textbf{MERaLiON2-Omni-r1}. This final step effectively ``tames'' the creative capabilities from step 1, and ensure that the reasoning process amplifies multimodal understanding rather than ``freely'' distorting it.

\section{Data Resources \& Benchmarks}
As stated above, we achieve the above systems progressively. In this section, we detail the data construction for the perception backbone (``System 1'') and the cognitive reasoning capabilities (``System 2''). Particularly, this section will also introduce the manually annotated SEA-Omni Benchmark Suite.

\subsection{Perception Backbone Training Data}
Our training of the perception backbone follows a three-stage protocol, where stages 1 and 2 are tailored for general and SEA visual alignment, while Stage 3 introduces a multimodal and multilingual expansion.

\begin{itemize}
    \item \textbf{Stage 1: Vision-Language Alignment.} We align the visual encoder with the LLM embedding space using around \textbf{158K} high-quality image-text and video-text pairs. The dataset aggregates open-source subsets from TextVQA~\citep{singh2019towards}, DocVQA~\citep{mathew2021docvqa}, MathVista~\citep{lu2023mathvista},
    DVQA~\citep{kafle2018dvqa}, WebVid-10M~\citep{bain2021frozen}, and Koala-36M-v1\footnote{\url{https://huggingface.co/datasets/Koala-36M/Koala-36M-v1}}, which are strictly filtered for visual grounding.
    \item \textbf{Stage 2: Deep Unimodal Visual Knowledge Ingestion.} We train on the \textbf{SEA-Image} corpus (with around 4.23M image-text pairs) to inject SEA visual knowledge. In general, this dataset establishes the vision encoder as a domain expert prior to multimodal fusion, covering Southeast Asian entities, foods, and landmarks in real scenarios.
    \item \textbf{Stage 3: Multilingual Multimodal Instruction Tuning.} This stage encompasses two kinds of data categories: \textit{Base Data.} We utilize the core multimodal instruction set (around \textbf{41.6K} samples) comprising \textbf{21.2K} multi-task images from SEA-VL, \textbf{3K} image-audio instances, \textbf{5.4K} video-audio instances, and a \textbf{12K} ASR subset. \textit{Multilingual Expansion.} To enforce cross-lingual ``omni'' capabilities, we translate the instruction-following portion of the base data into six regional languages: Chinese, Malay, Indonesian, Thai, Tamil, and Vietnamese. Thus, the model is finally fine-tuned on this mixed corpus to achieve instruction following across the region's linguistic landscape.
\end{itemize}

\subsection{Reasoning Instruction Set}
To obtain MERaLiON2-Omni-r1, we introduce a two-stage reasoning pipeline that filters prioritized reasoning patterns and adapts them to multimodal settings.

\begin{itemize}
    \item \textbf{Textual Logic Distillation.} We carefully formed a textual reasoning training corpus by filtering and ranking samples from the \textbf{AM-DeepSeek-R1-Distilled-1.4M} dataset~\citep{zhao202514millionopensourcedistilled}. We selected the top 423K samples exhibiting clear Chain-of-Thought (CoT) structures and verifiable answers, prioritizing math and logic to warm up the reasoning engine.
    \item \textbf{Multimodal Reasoning Alignment.} To ground reasoning in sensory inputs, we developed an in-house multimodal reasoning dataset comprising 21.3K samples. This dataset encompasses three categories: pure modality reasoning on standalone audio, text, or image inputs; mixed-modality reasoning that interleaves text, audio, and images for cross-referencing; and video reasoning focused on temporal logic in text–video pairs.
\end{itemize}

\subsection{The SEA-Omni Benchmark Suite}
To our knowledge, existing general-purpose benchmarks often fail to measure a model’s cultural grounding~\citep{liu2025culturevlm, kadiyala2025uncovering}. In this work, we introduce the \textbf{SEA-Omni Benchmark Suite}, consisting of separate image and video components, to assess our models' performance on SEA images and videos.
\begin{itemize}
    \item \textbf{SEA-Image Benchmark:} Built on the culturally diverse SEA-VL repository, the SEA-Image Benchmark is a large-scale dataset designed to evaluate vision–language alignment in Southeast Asian settings. The dataset is created through a semi-automated pipeline that combines a dedicated SEA visual analyzer for candidate generation with refinement from a teacher LLM, and the resulting annotations show high inter-rater agreement (ICC2 = 0.915). The benchmark covers five tasks with 2,049 samples on captioning, 2,049 on summarization, 1,199 on open-ended QA, 1,885 on OCR, and 1,175 on multiple-choice QA. To promote a realistic test of generalization, we enforce strict source separation, where training data comes from the web-crawled split, while evaluation uses exclusively crowdsourced samples.
    \item \textbf{SEA-Video Benchmark:} The SEA-Video Benchmark extends regional evaluation to dynamic settings through a manually curated dataset with 3,142 task videos. The benchmark highlights linguistic diversity across multiple languages including English, Mandarin, Indonesian, Thai, and Malay. Our annotations follow a Human–AI–Human workflow: domain experts first produce video summaries, which are used to prompt GPT-4o to generate culturally grounded multimodal questions. Then the questions are manually reviewed, refined, and answered by human annotators. By requiring joint reasoning over visual, audio, and textual signals, the benchmark focuses on perceptual reasoning rather than parametric recall, as reflected in a marked performance gap between text-only and multimodal baselines.
\end{itemize}

\section{Experiments}
\begin{table*}[t!]
\centering
\small
\begin{tabular}{l ccc c c ccc}
\toprule
\multirow{2}{*}{\textbf{Method}} & \textbf{Captioning} & \multicolumn{2}{c}{\textbf{QA}} & \textbf{Summ} & \textbf{MCQA} & \multicolumn{3}{c}{\textbf{OCR}} \\
\cmidrule(lr){2-2} \cmidrule(lr){3-4} \cmidrule(lr){5-5} \cmidrule(lr){6-6} \cmidrule(lr){7-9}
 & BERTScore & BERTScore & F1 & BERTScore & Acc & CER & WER & WER* \\
\midrule
SeaLion4VL-27B (2025)    & 77.64 & 85.16 & 14.71 & 82.92 & 83.32 & 0.40 & 0.45 & 0.094 \\
Qwen3Omni-30B             & 78.90 & 84.63 & 14.51 & 82.86 & 86.21 & 0.29 & 0.31 & 0.102\\
Qwen2.5Omni-7B           & 80.63 & 83.58 & 12.79 & 81.86 & 72.26 & 0.36 & 0.41 & 0.237 \\
\midrule
M2-Omni-Base      & 80.41 & 83.28 & 12.93 & 81.54 & 72.32 & 0.98 & 1.03 & 0.675 \\
M2-Omni-SEA   & 83.32 & 83.19 & 12.93 & 83.60 & 85.11 & 0.11 & 0.19 & 0.129 \\
M2-Omni-Instruct      & \underline{84.98} & \textbf{85.65} & \textbf{18.78} & \underline{84.13} & \underline{88.68} & \textbf{0.07} & \textbf{0.12} & \textbf{0.071} \\
M2-Omni-ML (\textbf{System 1}) & \textbf{85.19} & \underline{85.51} & \underline{18.27} & 83.64 & \textbf{91.06} & \textbf{0.07} & \textbf{0.12} & \underline{0.084} \\
M2-Omni-ML-r1 (\textbf{System 2}) & 82.43 & 85.04 & 14.94 & \textbf{84.36} & 78.98 & 0.17 & 0.18 & 0.140 \\
\bottomrule
\end{tabular}%
\caption{Performance on the SEA-Image Benchmark. In WER*, we ignore insertions to handle verbose outputs. The best results are in bold while the second best are underlined.}
\label{tab:table1}
\end{table*}

\begin{table}[t!]
\centering
\small
\setlength{\tabcolsep}{4.5pt}
\begin{tabular}{l cccccc}
\toprule
\multirow{2}{*}{\textbf{Method}} & \textbf{Summ} & \multicolumn{2}{c}{\textbf{AQA}} & \multicolumn{2}{c}{\textbf{VQA}} & \textbf{MCQA} \\
\cmidrule(lr){2-2} \cmidrule(lr){3-4} \cmidrule(lr){5-6} \cmidrule(lr){7-7}
 & BERTScore & BERTScore & F1 & BERTScore & F1 & Acc \\
\midrule
Qwen3Omni-30B   & 84.75 & 86.19 & 18.75 & 85.50 & 17.02 & 83.07 \\
Qwen2.5Omni-7B & 85.13 & 85.77 & 16.82 & 85.68 & 16.50 & 49.83 \\
\midrule
M2-Omni-Base      & 85.22 & 86.05 & 17.99 & 85.92 & 17.07 & 24.35 \\
M2-Omni-SEA   & 77.32 & 82.25 & 11.33 & 82.08 & 11.67 & 10.19 \\
M2-Omni-Instruct      & 85.10 & 87.50 & \underline{24.10} & 87.28 & \underline{23.37} & \underline{84.89} \\
M2-Omni-ML (\textbf{System 1})  & \underline{86.22} & \textbf{88.70} & \textbf{25.01} & \textbf{89.08} & \textbf{26.54} & \textbf{87.48} \\
M2-Omni-ML-r1 (\textbf{System 2})  & \textbf{87.07} & \underline{87.80} & 22.21 & \underline{87.93} & 22.45 & 81.61 \\
\bottomrule
\end{tabular}
\caption{Performance on the SEA-Video Benchmark.}
\label{tab:table2}
\end{table}

\subsection{Main Results}
We evaluate MERaLiON2-Omni across various capability axes including general perception, textual reasoning, spoken cognition, among others.

\paragraph{Perception \& Regional Alignment (System 1 Validation).}
Tables~\ref{tab:table1} and~\ref{tab:table2} show the regional superiority of our proposed Perception Backbone M2-Omni-ML model on SEA-Omni benchmarks compared with Qwen2.5Omni, Qwen3Omni, and SeaLion4VL\footnote{\textbf{Gemma-SEA-LION-v4-27B-VL} is a vision-language model derived from Gemma-SEA-LION-v4-27B-IT by supervised post-training on instruction-image pairs in multiple Southeast Asian languages. It inherits the large context window and multimodal image-and-text understanding capabilities from the underlying Gemma 3 architecture.}.
Although we observe a slight multilingual ``tax'' on English-centric benchmarks in Table~\ref{tab:table3}, we deem this trade-off highly acceptable given the model's expanded regional utility.

As shown in Table~\ref{tab:table1}, the backbone achieves competitive performance in OCR tasks (WER* 0.084)\footnote{\textbf{WER*} is defined as $(S + D) / N$ to measure substitutions and deletions while ignore insertions. This adjusted WER serves as a reference metric to prevent the penalization of valid, albeit verbose, explanatory content generated by the model during open-ended cultural reasoning.} and Visual MCQA (Acc 91.06). Interestingly, the reasoning-enhanced model (r1) sees a notable regression in these ``pure perception'' tasks (OCR degraded to 0.140-WER*). This suggests that while reasoning helps abstract logic, it may introduce ``over-thinking'' noise into low-level character recognition, where direct visual mapping could be preferred over logical deduction.

\begin{table*}[t!]
\centering
\small
\begin{tabular}{l ccc ccc}
\toprule
\multirow{2}{*}{\textbf{Model}} & \multicolumn{3}{c}{\textbf{Video Benchmarks}} & \multicolumn{3}{c}{\textbf{Image Benchmarks}} \\
\cmidrule(lr){2-4} \cmidrule(lr){5-7}
 & Video-MME & Video-MME-sub & MVBench & MME & MMB-V11 & MMVet \\
\midrule
Qwen3Omni-30B         & 61.48 & 66.22 & 52.11 & 2,043 & 86.18 & 73.90 \\
Qwen2.5Omni-7B        & 54.19 & 58.63 & 53.96 & 1,756 & 81.97 & 61.10 \\
\midrule
M2-Omni-Instruct & 52.52 & 56.67 & 36.98 & 1,669 & 62.03 & 39.30 \\
M2-Omni-ML & 50.26 & 56.00 & 37.08 & 1,667 & 58.02 & 36.40 \\
\bottomrule
\end{tabular}%
\caption{Performance comparison on 6 general-purpose multimodal benchmarks~\citep{zhang2021mme, li2024mvbench, yu2024mm}. It shows a minor performance drop when expanding from the English-centric checkpoint to the multilingual backbone.}
\label{tab:table3}
\end{table*}
\begin{table}[t]
 \small
 \centering
 \setlength{\tabcolsep}{4.5pt}
\begin{tabular}{lcc}
\toprule
\textbf{Dataset (cases)} & \textbf{M2O-ML (System 1)} & \textbf{M2O-ML-r1 (System 2)} \\
\midrule
AM\_evol-zh (50) & 0.2800 & \textbf{0.3000} \\
AM\_NuminaMath\_1.5 (197) & 0.2589 & \textbf{0.3706} \\
AM\_InfinityInstruct (1029) & 0.4451 & \textbf{0.4898} \\
AM\_natural\_reasoning (1101) & 0.4087 & \textbf{0.4959} \\
AM\_OpenCoderStage2 (78) & 0.3718 & \textbf{0.3974} \\
AM\_PRIME (31) & \textbf{0.2258} & 0.1935 \\
AM\_None (43) & 0.3721 & \textbf{0.5116} \\
AM\_codeio (191) & \textbf{0.2513} & 0.2461 \\
AM\_prime (34) & 0.1471 & \textbf{0.1765} \\
AM\_evol-en (61) & \textbf{0.4262} & 0.3934 \\
AM\_OpenCoder (67) & \textbf{0.6119} & 0.4030 \\
AM\_MATH\_metamathQA (50) & 0.8000 & \textbf{0.8400} \\
AM\_open\_orca (34) & \textbf{0.8235} & \textbf{0.8235} \\
AM\_Others (34) & 0.2353 & \textbf{0.3529} \\
\bottomrule
\end{tabular}
\caption{Performance comparison between M2-Omni-ML and M2-Omni-ML-r1 across evaluation datasets. Results are reported as Consistency Scores (0-1) evaluated by a Qwen3-235B Judge against ground-truth. Best result per dataset is shown in bold.}
\label{tab:table4}
\end{table}

\paragraph{The ``Cognitive Boost'' in Text (System 2 Validation).}
To verify the efficacy of our reasoning training, we evaluate the text-only performance of the backbone versus the reasoning-enhanced M2-Omni-ML-r1. As detailed in Table~\ref{tab:table4}, the r1 model achieves significant gains on abstract logic tasks. Specifically, the performance on \textbf{NuminaMath} increases by \textbf{+11.2\%} (from 0.2589 to 0.3706).
Besides, we also observe a ``Rigidity Penalty'' in strict syntax tasks, where the performance of \textbf{OpenCoder} drops significantly (-21 points). Our analysis suggests that CoT training encourages exploration, but this can sometimes hurt the model’s ability to memorize exact code syntax. In other words, there is a trade-off between creativity and control where more flexible reasoning may lead to less precise outputs.

We also observe that this ``Rigidity Penalty'' not only affects syntax-heavy tasks, it also appears in open-ended multimodal QA, such as SEA-Video (see Table~\ref{tab:table2}). However, our case study shows that the M2O-ML-r1 model sometimes fails these metrics because it gives too much detail, including culturally nuanced explanations that the ground truth doesn’t have. We believe that this single-objective and strictly matched evaluation method is not fully compatible with the goals of full-modal reasoning, and further research is needed on how to better evaluate such models.
\begin{table*}[t!]
\centering
\small
\resizebox{\textwidth}{!}{%
\begin{tabular}{l ccccrc | c}
\toprule
\textbf{Dataset} & \textbf{M2-AL} & \textbf{M2O-Inst} & \textbf{M2O-ML} & \textbf{M2O-ML-r1} & \textbf{M2O-ML-r1*} & \textbf{Qwen2.5Omni} & \textbf{Qwen3Omni*} \\
\midrule
\multicolumn{7}{l}{\textit{\textbf{English Datasets}}} \\
Librispeech\_test\_clean      & \underline{0.025} & \textbf{0.024} & 0.026 & 0.044 & 0.040 (0.4\%) \textcolor{blue}{\textbf{$\uparrow$}} & 0.134 & 0.016 \\
Librispeech\_test\_other      & \textbf{0.043} & \textbf{0.043} & 0.046 & 0.051 & 0.051 (0.2\%) - & 0.164 & 0.026 \\
Commonvoice\_15\_en\_test           & \textbf{0.084} & \textbf{0.084} & 0.092 & 0.092 & 0.091 (0.1\%) \textcolor{blue}{\textbf{$\uparrow$}} & 0.290 & 0.061 \\
Peoples\_speech\_test              & \textbf{0.195} & \underline{0.207} & 0.226 & 0.249 & 0.247 (0.6\%) \textcolor{blue}{\textbf{$\uparrow$}} & 0.491 & 0.153 \\
Gigaspeech\_test                  & \textbf{0.094} & \underline{0.097} & 0.105 & 0.100 & 0.098 (0.5\%) \textcolor{blue}{\textbf{$\uparrow$}} & 0.318 & 0.088 \\
Tedlium3\_test                    & \textbf{0.036} & \underline{0.038} & 0.039 & 0.041 & 0.037 (0.3\%) \textcolor{blue}{\textbf{$\uparrow$}} & 0.243 & 0.025 \\
Tedlium3\_long\_form\_test          & \textbf{0.669} & 0.736 & \underline{0.684} & 0.727 & 0.727 (0.0\%) - & 0.900 & 0.967 \\
Earnings21\_test                  & \underline{0.917} & 0.925 & 0.941 & \textbf{0.913} & 0.913 (0.0\%) - & 0.937 & 0.991 \\
Earnings22\_test                  & \underline{0.927} & 0.937 & 0.951 & \textbf{0.925} & 0.925 (0.0\%) - & 0.943 & 0.992 \\
\midrule
\multicolumn{7}{l}{\textit{\textbf{Multilingual}}} \\
Commonvoice\_17\_id\_asr (Indonesian)        & 0.111 & \underline{0.088} & \textbf{0.087} & 0.487 & 0.128 (38.2\%) \textcolor{blue}{\textbf{$\uparrow$}} & 0.829 & 0.098 \\
Commonvoice\_17\_ta\_asr (Tamil)             & \underline{0.158} & 0.161 & \textbf{0.155} & 0.368 & 0.236 (21.8\%) \textcolor{blue}{\textbf{$\uparrow$}} & 1.016 & 0.421 \\
Commonvoice\_17\_th\_asr (Thai)              & \underline{0.387} & 0.426 & \textbf{0.329} & 0.440 & 0.128 (35.8\%)\textcolor{blue}{\textbf{$\uparrow$}} & 0.935 & 0.034 \\
Commonvoice\_17\_vi\_asr (Vietnamese)        & \textbf{0.157} & \underline{0.164} & 0.177 & 0.336 & 0.188 (27.8\%) \textcolor{blue}{\textbf{$\uparrow$}} & 0.780 & 0.117 \\
Commonvoice\_zh\_asr (Chinese)             & \underline{0.168} & \textbf{0.147} & 0.173 & 0.364 & 0.351 (7.2\%) \textcolor{blue}{\textbf{$\uparrow$}} & 0.622 & 0.096 \\
Commonvoice\_ss\_2.0 (Bahasa Malay)          & \textbf{0.094} & \underline{0.109} & 0.112 & 0.110 & 0.097 (9.9\%)\textcolor{blue}{\textbf{$\uparrow$}} & 0.436 & 0.260 \\
Aishell\_asr\_zh\_test (Chinese)         & \underline{0.067} & 0.073 & \textbf{0.061} & 0.108 & 0.104 (1.4\%) \textcolor{blue}{\textbf{$\uparrow$}} & 0.698 & 0.056 \\
Fleurs\_tamil\_ta\_30\_asr (Tamil)           & \textbf{0.169} & 0.179 & \underline{0.171} & 0.488 & 0.277 (36.0\%) \textcolor{blue}{\textbf{$\uparrow$}} & 1.065 & 0.346 \\
Lotus\_thai\_th\_30\_asr (Thai)             & \underline{0.021} & \textbf{0.020} & 0.025 & 0.123 & 0.025 (13.8\%) \textcolor{blue}{\textbf{$\uparrow$}} & 0.928 & 0.012 \\
\midrule
\multicolumn{7}{l}{\textit{\textbf{Code-Switching}}} \\
Seame\_dev\_man               & 0.193 & \textbf{0.169} & \underline{0.180} & 0.475 & 0.461 (14.6\%) \textcolor{blue}{\textbf{$\uparrow$}} & 1.332 & 0.439 \\
Seame\_dev\_sge               & 0.204 & \underline{0.202} & \textbf{0.201} & 0.829 & 0.887 (7.9\%) \textcolor{red}{$\downarrow$} & 1.255 & 0.481 \\
\bottomrule
\end{tabular}
}
\caption{WER performance comparison on English, Multilingual (SEA), and Code-Switching datasets. M2O-ML-r1* refers to an ablation study where we count the number of ASR results that are translated into the reasoning language, and we report the performance with these language-drift generation removed to reveal their effects. Qwen3Omni* (30B) is included only for reference.}
\label{tab:final_results}
\end{table*}

\paragraph{SEA ASR \& Code-Switching Robustness.}
Table~\ref{tab:final_results} presents the performance of our models across 19 ASR benchmarks. A key question we investigate is whether integrating the visual modality causes catastrophic degradation of existing audio capabilities. The results in Table~\ref{tab:final_results} show that our M2O-Instruct and M2O-ML models largely preserve the auditory ability of our Audio-LLM (M2-AL) and even yield modest performance gains in certain scenarios. For example, on the \textit{Seame\_dev\_man} code-switching dataset, M2O-Instruct improves recognition accuracy with WER score reduced from 0.193 (M2-AL) to 0.169. This validates our finding in Section 3.1 that the newly learned visual ability is ``geometrically orthogonal'' to the auditory ability, reducing cross-modality interference.

While the 30B Qwen3Omni model performs strongly on high-resource languages (e.g., Chinese and English), it exhibits substantially higher WER on regional or code-switched data (e.g., 0.439 on \textit{Seame\_dev\_man}), whereas our 10B models maintain comparatively lower WER scores (e.g., 0.169 of M2O-Instruct). Notably, on long-form financial audio benchmarks (e.g., \textit{Tedlium3\_long\_form\_test}, \textit{Earnings21\_test}, and \textit{Earnings22\_test}), our models perform consistently better.

However, as shown in the M2O-ML-r1 column, the injection of CoT capabilities leads to a notable regression in ASR (e.g., \textit{Seame\_dev\_man} WER rises to 0.475).
To investigate this phenomenon, we analyze the model's ASR results and find that r1 tends to translate the speech content using its reasoning language, leading to high WER scores in many datasets. To intuitively reveal it, we conducted an ablation study, where we exclude those translated instances for evaluation. The results of \textbf{M2O-ML-r1*} indicate a severe effect of this language drifting tendency on ASR. In general, the r1 model (System-2) outperforms M2O-ML (System-1) on 8/20 datasets and remain competitive on others.
Besides, we argue that System-2 may prioritize semantic interpretation over word-for-word transcription, which could affect traditional ASR fidelity. This phenomenon implies a fundamental trade-off between semantic reasoning and perception.

\begin{table*}[t!]
\centering
\small
\resizebox{\textwidth}{!}{%
\begin{tabular}{l l cccc c | c}
\toprule
\multirow{2}{*}{\textbf{Channel}} & \multirow{2}{*}{\textbf{Dataset}} & \multicolumn{4}{c}{\textbf{MERaLiON Variants}} & \textbf{Baseline} & \textbf{Reference} \\
\cmidrule(lr){3-6} \cmidrule(lr){7-7} \cmidrule(lr){8-8}
 &  & M2-AL & M2O-Inst & M2O-ML & M2O-ML-r1 & Qwen2.5Omni & Qwen3Omni* \\
\midrule

\multirow{5}{*}{\shortstack[l]{Spoken QA\\(English)}}
 & cn\_college\_listen\_mcq & 76.50 & 75.80 & 78.50 & \textbf{88.80} \textcolor{blue}{\textbf{$\uparrow$}} & 78.70 & 93.90 \\
 & slue\_p2\_sqa5 & 89.51 & 90.34 & \textbf{92.45} & 90.39 \textcolor{red}{$\downarrow$} & 82.55 & 91.42 \\
 & dream\_tts\_mcq & 71.90 & 74.80 & 77.05 & \textbf{85.70} \textcolor{blue}{\textbf{$\uparrow$}} & 72.00 & 87.10 \\
 & public\_sg\_speech\_qa & 73.28 & 78.90 & \textbf{79.80} & 78.90 \textcolor{red}{$\downarrow$} & 65.87 & 80.96 \\
 & spoken\_squad & 89.04 & 87.70 & \textbf{90.96} & 88.42 \textcolor{red}{$\downarrow$} & 77.46 & 82.24 \\
\midrule

\multirow{4}{*}{\shortstack[l]{Spoken QA\\(Singlish)}}
 & imda\_part3\_30s\_sqa & 58.40 & 60.60 & 57.20 & \textbf{61.40} \textcolor{blue}{\textbf{$\uparrow$}} & 45.00 & 66.20 \\
 & imda\_part4\_30s\_sqa & 58.80 & 61.20 & 61.20 & \textbf{64.00} \textcolor{blue}{\textbf{$\uparrow$}} & 47.60 & 69.20 \\
 & imda\_part5\_30s\_sqa & 68.40 & 72.40 & 70.40 & \textbf{73.20} \textcolor{blue}{\textbf{$\uparrow$}} & 60.40 & 78.00 \\
 & imda\_part6\_30s\_sqa & 70.80 & 73.80 & \textbf{75.60} & 75.40 \textcolor{red}{$\downarrow$} & 63.00 & 76.40 \\
\midrule

\multirow{4}{*}{\shortstack[l]{Spoken Dialogue\\Summarization}}
 & imda\_part3\_30s\_ds & 51.20 & \textbf{52.00} & 48.00 & 49.40 \textcolor{blue}{\textbf{$\uparrow$}} & 35.40 & 57.40 \\
 & imda\_part4\_30s\_ds & 49.60 & 48.80 & \textbf{51.00} & 48.60 \textcolor{red}{$\downarrow$} & 29.20 & 57.00 \\
 & imda\_part5\_30s\_ds & 57.80 & 57.60 & 55.20 & \textbf{60.00} \textcolor{blue}{\textbf{$\uparrow$}} & 47.00 & 68.60 \\
 & imda\_part6\_30s\_ds & 61.40 & 65.00 & 63.00 & \textbf{66.60} \textcolor{blue}{\textbf{$\uparrow$}} & 51.20 & 71.00 \\
\midrule

\multirow{4}{*}{\shortstack[l]{Gender\\Recognition}}
 & voxceleb\_gender & 97.30 & 99.30 & \textbf{99.50} & 89.30 \textcolor{red}{$\downarrow$} & 17.60 & 97.80 \\
 & iemocap\_gender & 94.80 & 96.30 & \textbf{97.50} & 74.10 \textcolor{red}{$\downarrow$} & 23.60 & 92.00 \\
 & imda\_gr\_sentence & 53.80 & 56.80 & \textbf{59.00} & 47.30 \textcolor{red}{$\downarrow$} & 1.80 & 88.90 \\
 & imda\_gr\_dialogue & 70.40 & 74.30 & \textbf{76.10} & 36.80 \textcolor{red}{$\downarrow$} & 3.70 & 81.30 \\
\midrule

\multirow{3}{*}{\shortstack[l]{Audio Scene\\QA}}
 & clotho\_aqa & 56.86 & \textbf{61.98} & 61.72 & 53.42 \textcolor{red}{$\downarrow$} & 42.04 & 69.10 \\
 & wavcaps\_qa & 44.54 & 45.99 & \textbf{49.87} & 40.79 \textcolor{red}{$\downarrow$} & 35.79 & 54.74 \\
 & audiocaps\_qa & 50.03 & 51.25 & \textbf{52.65} & 42.11 \textcolor{red}{$\downarrow$} & 40.64 & 59.62 \\
\midrule

\multirow{3}{*}{\shortstack[l]{Emotion\\Recognition}}
 & iemocap\_emotion & 61.00 & 59.60 & \textbf{65.10} & 42.60 \textcolor{red}{$\downarrow$} & 42.60 & 61.80 \\
 & meld\_sentiment & \textbf{64.50} & 47.60 & 50.60 & 53.00 \textcolor{blue}{\textbf{$\uparrow$}} & 33.40 & 58.10 \\
 & meld\_emotion & \textbf{54.50} & 40.70 & 40.20 & 47.50 \textcolor{blue}{\textbf{$\uparrow$}} & 35.60 & 59.30 \\
\midrule

\multirow{6}{*}{\shortstack[l]{Speech-\\Instruct}}
 & openhermes\_audio & 60.80 & 58.20 & 58.80 & \textbf{69.20} \textcolor{blue}{\textbf{$\uparrow$}} & 64.00 & 82.00 \\
 & alpaca\_audio & 72.00 & 65.60 & 63.00 & \textbf{75.00} \textcolor{blue}{\textbf{$\uparrow$}} & 67.60 & 82.60 \\
 & spoken-mqa\_short & \textbf{0.90} & 0.67 & 0.71 & \phantom{0}0.64 \textcolor{red}{$\downarrow$} & \textbf{0.90} & 0.93 \\
 & spoken-mqa\_long & \textbf{0.50} & 0.35 & 0.35 & \phantom{0}0.41 \textcolor{blue}{\textbf{$\uparrow$}} & 0.48 & 0.88 \\
 & spoken-mqa\_1-step & 0.68 & 0.63 & 0.84 & \phantom{0}\textbf{0.87} \textcolor{blue}{\textbf{$\uparrow$}} & \textbf{0.87} & 0.88 \\
 & spoken-mqa\_m-step & 0.36 & 0.28 & 0.73 & \phantom{0}\textbf{0.79} \textcolor{blue}{\textbf{$\uparrow$}} & 0.77 & 0.81 \\
\midrule

\multirow{6}{*}{\shortstack[l]{Speech\\Translation}}
 & covost2\_en\_id & \textbf{36.89} & 28.04 & 30.22 & 32.57 \textcolor{blue}{\textbf{$\uparrow$}} & 6.69 & 28.64 \\
 & covost2\_en\_zh & \textbf{31.16} & 17.20 & 18.65 & 25.03 \textcolor{blue}{\textbf{$\uparrow$}} & 11.54 & 27.42 \\
 & covost2\_en\_ta & \textbf{21.44} & 11.68 & 12.81 & 16.83 \textcolor{blue}{\textbf{$\uparrow$}} & 0.47 & 22.48 \\
 & covost2\_id\_en & \textbf{46.32} & 28.46 & 29.24 & 36.20 \textcolor{blue}{\textbf{$\uparrow$}} & 10.20 & 42.35 \\
 & covost2\_zh\_en & \textbf{20.18} & 14.94 & 17.10 & 17.15 \textcolor{blue}{\textbf{$\uparrow$}} & 10.76 & 26.98 \\
 & covost2\_ta\_en & \textbf{3.16} & 1.57 & 1.58 & \phantom{0}2.13 \textcolor{blue}{\textbf{$\uparrow$}} & 0.05 & 0.60 \\

\bottomrule
\end{tabular}%
}
\caption{Detailed performance comparison on the standard AudioBench tasks~\citep{wang2024audiobench}. Colored arrows in the \textbf{M2O-ML-r1} column indicate performance gain (\textcolor{blue}{\textbf{$\uparrow$}}) or loss (\textcolor{red}{$\downarrow$}) compared to the \textbf{M2O-ML} model. We mainly compare our models against the comparable baseline \textbf{Qwen2.5Omni} (7B). \textbf{Qwen3Omni*} (30B) results are provided for reference only and are not included in the best-result comparison (bolded).}
\label{tab:table5}
\end{table*}

\begin{table*}[t!]
\centering
\resizebox{\textwidth}{!}{%
\begin{tabular}{l cc cc cc cc cc cc}
\toprule
\multirow{2}{*}{\textbf{Model}} & \multicolumn{2}{c}{\textbf{PB-Read}} & \multicolumn{2}{c}{\textbf{Topic-Read}} & \multicolumn{2}{c}{\textbf{Daily Conv.}} & \multicolumn{2}{c}{\textbf{Code-Switch}} & \multicolumn{2}{c}{\textbf{Themed Conv.}} & \multicolumn{2}{c}{\textbf{Service Call}} \\
\cmidrule(lr){2-3} \cmidrule(lr){4-5} \cmidrule(lr){6-7} \cmidrule(lr){8-9} \cmidrule(lr){10-11} \cmidrule(lr){12-13}
 & WER & WER* & WER & WER* & WER & WER* & WER & WER* & WER & WER* & WER & WER* \\
\midrule
Qwen3Omni-30B & 0.062 & 0.053 & 0.324 & 0.242 & \underline{0.278} & 0.221 & 0.440 & 0.368 & \textbf{0.187} & \underline{0.151} & \textbf{0.147} & \underline{0.122} \\
Qwen2.5Omni-7B & 0.291 & 0.237 & 0.792 & 0.598 & 0.820 & 0.744 & 0.886 & 0.851 & 0.751  & 0.655 & 0.783 & 0.666 \\
M2-AL~\citep{he2025meralion} & 0.059 & 0.048 & 0.144 & 0.130 & \textbf{0.263} & \underline{0.210} & \underline{0.342} & \textbf{0.268} & \underline{0.199} & 0.154 & 0.160 & 0.123 \\
M2O-Base & 0.108 & 0.081 & 0.262 & 0.234 & 0.559 & 0.490 & 0.696 & 0.648 & 0.504 & 0.444 & 0.426 & 0.348 \\
M2O-Instruct & \textbf{0.055} & \underline{0.046} & \textbf{0.112} & \textbf{0.101} & 0.284 & 0.239 & \textbf{0.338} & 0.279 & \underline{0.199} & 0.160 & \underline{0.156} & 0.128 \\
M2O-ML (\textbf{System 1}) & \underline{0.056} & \textbf{0.045} & \textbf{0.112} & \textbf{0.101} & 0.294 & 0.242 & 0.367 & 0.292 & 0.206 & 0.161 & 0.165 & 0.130 \\
M2O-ML-r1 (\textbf{System 2}) & 0.061\textcolor{red}{$\downarrow$} & \underline{0.046}\textcolor{red}{$\downarrow$} & 0.172\textcolor{red}{$\downarrow$} & 0.135\textcolor{red}{$\downarrow$} & 0.346\textcolor{red}{$\downarrow$} & \textbf{0.201}\textcolor{blue}{\textbf{$\uparrow$}} & 0.586\textcolor{red}{$\downarrow$} & \underline{0.273}\textcolor{blue}{\textbf{$\uparrow$}} & 0.376\textcolor{red}{$\downarrow$} & \textbf{0.142}\textcolor{blue}{\textbf{$\uparrow$}} & 0.254\textcolor{red}{$\downarrow$} & \textbf{0.107}\textcolor{blue}{\textbf{$\uparrow$}} \\
\bottomrule
\end{tabular}
}
\caption{Performance comparison of different models on six Singlish-ASR datasets~\citep{wang2025advancing}.}
\label{tab:table6}
\end{table*}

\paragraph{Audio-Visual Reasoning Trade-offs.}
Tables~\ref{tab:table5} and~\ref{tab:table6} present performance of our models on several audio tasks. The overall results indicate a clear trade-off, that is, although our injected reasoning capability significantly benefits high-level understanding (e.g., interpreting spoken instructions), it degrades performance on perceptual tasks like audio recognition.

\begin{itemize}
  \item In tasks requiring multi-step reasoning, our reasoning model shows clear performance gains. For example, M2O-ML-r1 reaches 88.80 on \textbf{Spoken QA (English)} compared to the backbone model’s 78.50. Besides, the r1 model scores 75.0 on \textbf{Speech Instruction (Alpaca Audio)}, suggesting that CoT abilities learned from text can be effectively transferred to audio instruction following.
  \item However, we observe that M2O-ML-r1's performance on perceptual tasks often degrades, as shown in Table~\ref{tab:table5}. For example, in Gender Recognition, the performance drops from 99.5\% to 89.3\% (Acc). The effect in Emotion Recognition looks more nuanced, we find that applying reasoning can improve performance on semantic-rich dialogues (Lines 22-23) but harm prosody-heavy tasks (Line 21). This indicates the reasoning model's tendency to focus more on logical content while ignoring the sound of the voice.
  \item In Table~\ref{tab:table6}, while M2O-ML-r1's WER score grows from 0.367 to 0.586, the adjusted WER* score actually drops to 0.273 when compared with the M2O-ML model. The reasoning model appears to edit and correct the transcription rather than reproduce it verbatim. In addition, even after post-processing, the model still favors logical coherence over faithful transcription, suggesting that without carefully designed prompts, reasoning may reduce the accuracy of speech transcription.
\end{itemize}

\subsection{Analysis: The Efficiency-Stability Paradox}

\begin{figure}[t]
  \centering
  \includegraphics[scale=0.34]{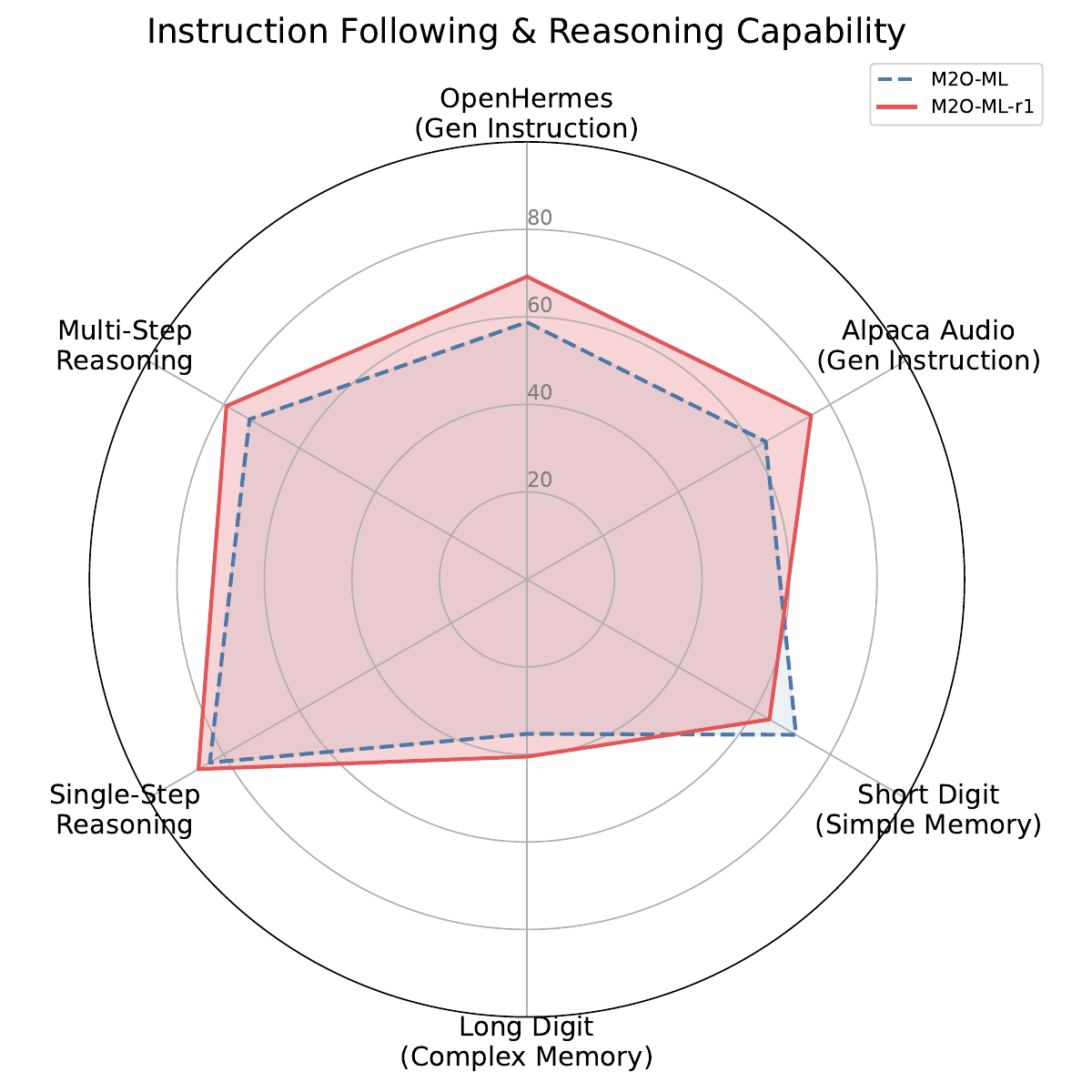}
  \caption{\textbf{The cognitive gap.} This radar chart illustrates the performance of our reasoning (M2O-ML-r1) and non-reasoning (M2O-ML) models on 6 representative AudioBench tasks.}
  \label{fig:1}
\end{figure}

\begin{figure}[t]
  \centering
  \includegraphics[scale=0.34]{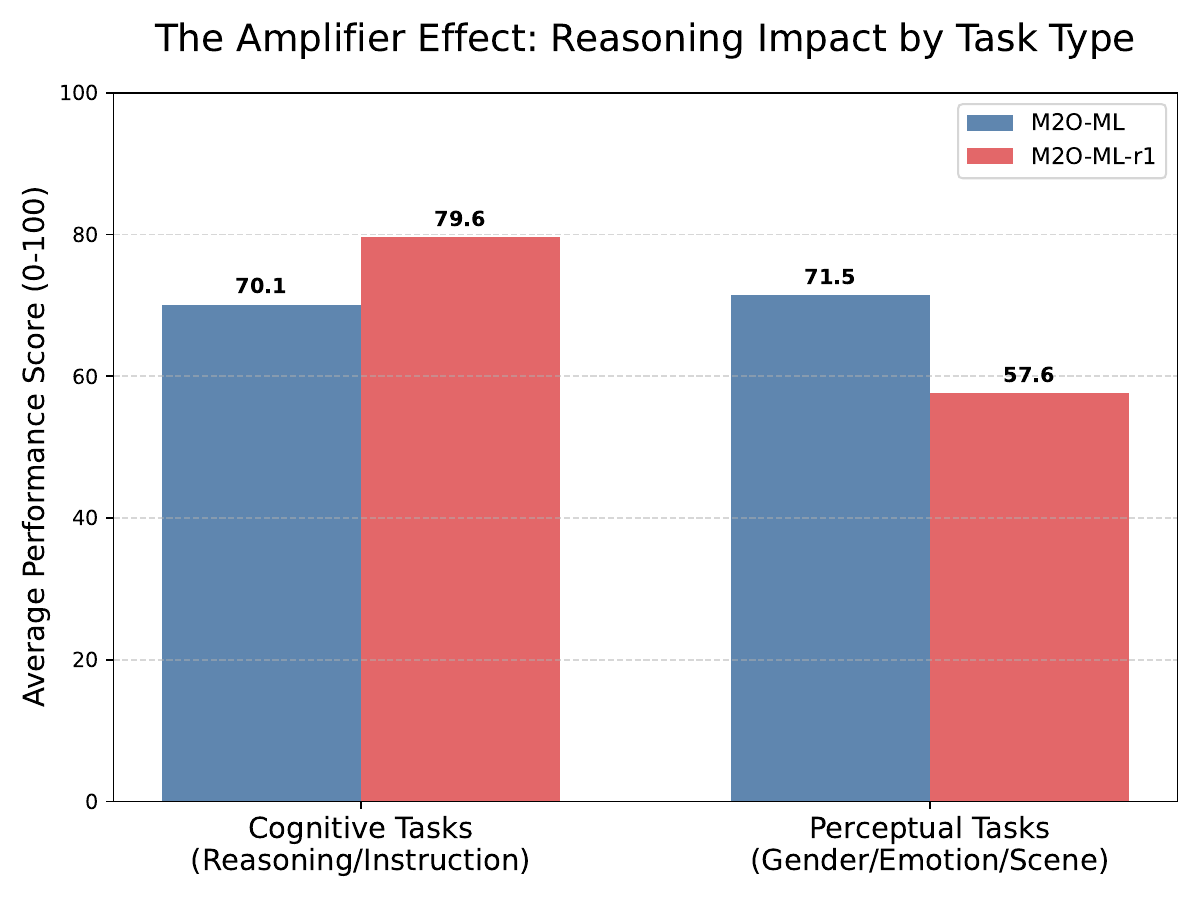}
  \caption{\textbf{The amplifier effect.} The introduced reasoning capability benefits cognitive tasks but harms perceptual tasks.}
  \label{fig:2}
\end{figure}

\subsubsection{Trade-off Analysis: Cognitive Gains vs. Perceptual Costs}
\textbf{Cognitive performance.} The radar chart in Figure~\ref{fig:1} shows that our reasoning model (M2O-ML-r1 colored in red) achieves stronger overall performance than the non-reasoning baseline. In addition, on complex tasks like Multi-Step Reasoning and Spoken Instruction following, M2O-ML-r1 consistently outperforms M2O-ML. We attribute this gap to the model’s ability to decompose instructions, which allows the model to handle more complicated planning and logical inference tasks.

\textbf{Perception trade-offs.} However, these performance gains come with a cost. As we can see in Figure~\ref{fig:2}, although M2O-ML-r1 boosts logic and instruction-following scores by a significant margin (the left two bars), it harms perceptual tasks like gender or emotion recognition (the right two). We interpret this decline as evidence of cognitive overload. In other words, applying ``System 2'' reasoning to tasks suited for ``System 1'' can overcomplicate signals, hurting transcription accuracy, particularly for models trained in low-resource settings.

\begin{figure}[t]
  \centering
  \includegraphics[scale=0.34]{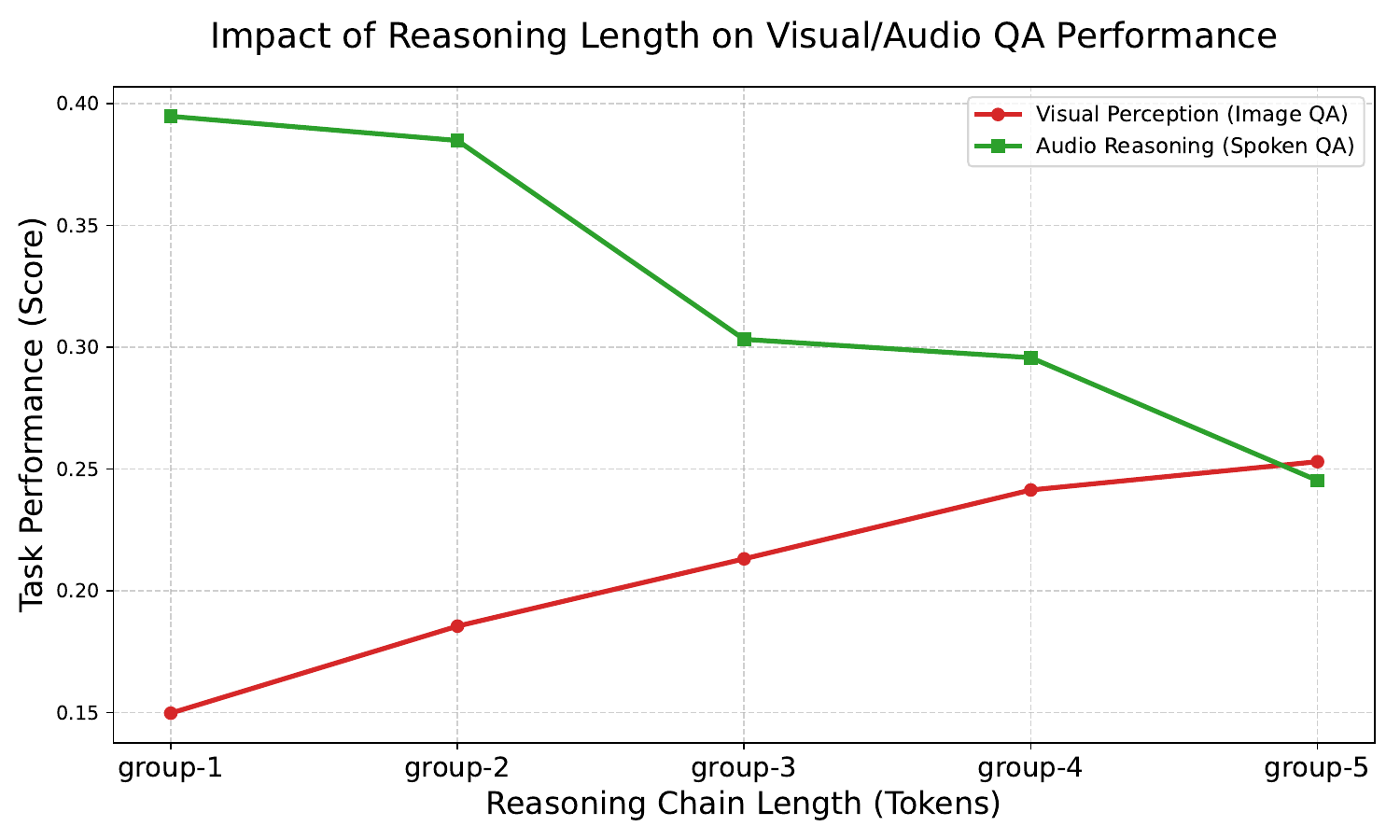}
  \caption{\textbf{Divergent impact of reasoning length.} Visual accuracy improves with longer reasoning (the red line), while Audio accuracy degrades due to Temporal Drift (the green line).}
  \label{fig:3}
\end{figure}
\begin{figure}[t]
  \centering
  \includegraphics[scale=0.34]{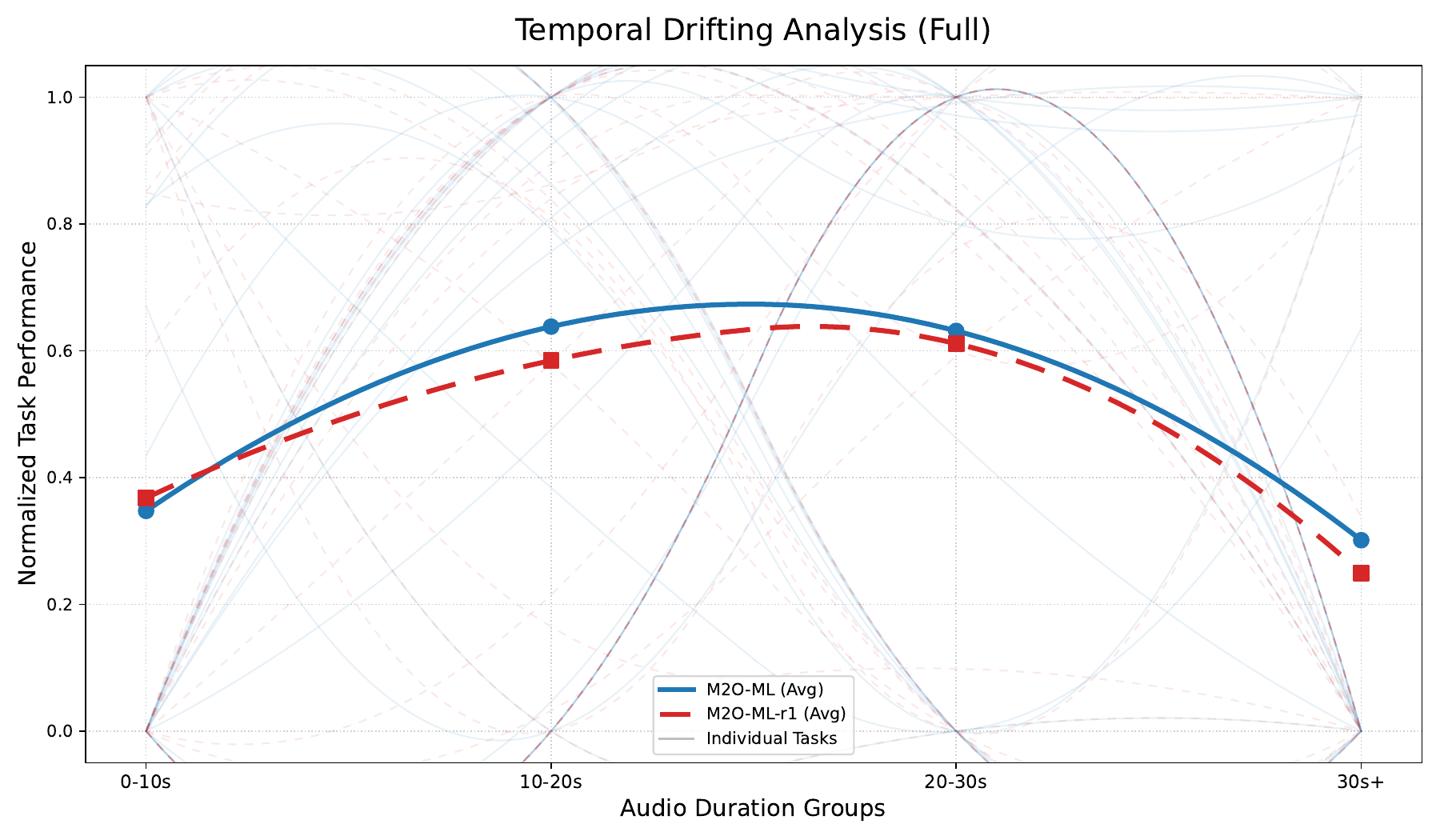}
  \caption{\textbf{The anatomy of temporal drift}.}
  \label{fig:5}
\end{figure}

\subsubsection{Visual Decompression vs. Temporal Drift}
In Figure~\ref{fig:3}, we further analyze the relationship between reasoning chain length and accuracy across different modalities to explore how time affects multi-modal processing.
\begin{itemize}
    \item \textbf{Visual Decompression (Upward Slope):} In static visual QA, we find that longer reasoning chains correlate with higher accuracy which we term as ``\textbf{Visual Decompression}.'' The model uses the extra ``thought tokens'' to explicitly map dense pixel information into the language space (e.g., “I see a red car... therefore...”) before concluding. Since the image is static and the ``evidence'' never expires, it allows the model to ``talk itself through'' the visual scene.
    \item \textbf{Temporal Drift (Downward Slope):} Conversely, we find that Audio QA exhibits a negative correlation. Unlike static images, audio is inherently temporal, however, as the model generates long CoT sequences, it consumes inference time that seemingly desynchronizes it from the audio buffer, which we term ``\textbf{Temporal Drift}.'' In this case, when the model completes its reasoning chain, it may have ``forgotten'' the specific acoustic timestamp it was referencing, therefore leading to a disconnect between the reasoning path and the audio clues.
\end{itemize}

In Figure~\ref{fig:5}, we take a closer look at how temporal drift varies with the duration of input audios:
\begin{itemize}
    \item For short audio clips (<10s), we find that our reasoning model M2O-ML-r1 consistently outperforms M2O-ML. We attribute this gain to the compact audio context, which lets extra reasoning steps work without overloading the model’s memory.
    \item However, once the audio exceeds roughly 30 seconds, M2O-ML-r1’s performance degrades a lot, falling behind the non-reasoning model. We attribute this primarily to the token overhead from CoT reasoning, which we suspect dilutes the attention given to acoustic features, thereby weakening temporal alignment.
\end{itemize}

\begin{figure}[t]
  \centering
  \includegraphics[scale=0.34]{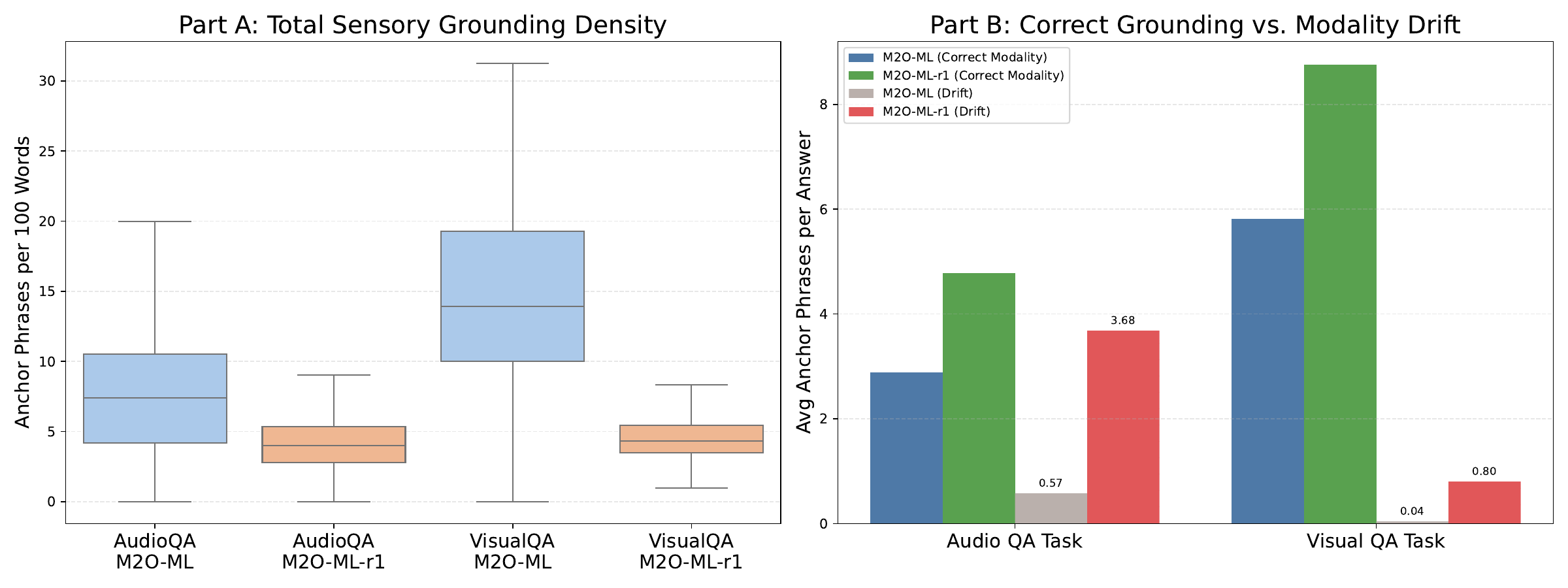}
  \caption{\textbf{Anchor token analysis.} In Audio QA tasks (where inputs comprise both audio and video frames), the reasoning model generates a significantly higher number of visual-anchor tokens (e.g., colors, shapes) compared to M2O-ML.}
  \label{fig:4}
\end{figure}

\begin{figure}[t]
  \centering
  \begin{subfigure}[t]{0.48\linewidth}
    \centering
    \includegraphics[width=\linewidth]{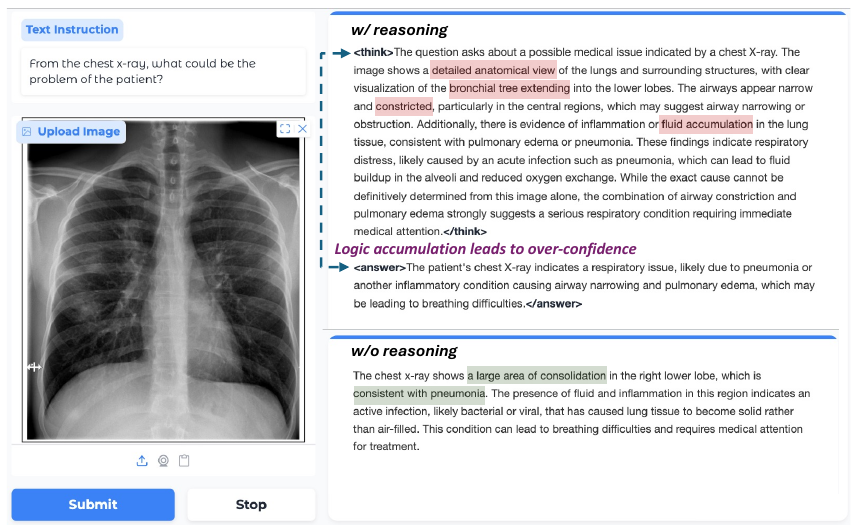}
    \caption{}
  \end{subfigure}
  \hfill
  \begin{subfigure}[t]{0.48\linewidth}
    \centering
    \includegraphics[width=\linewidth]{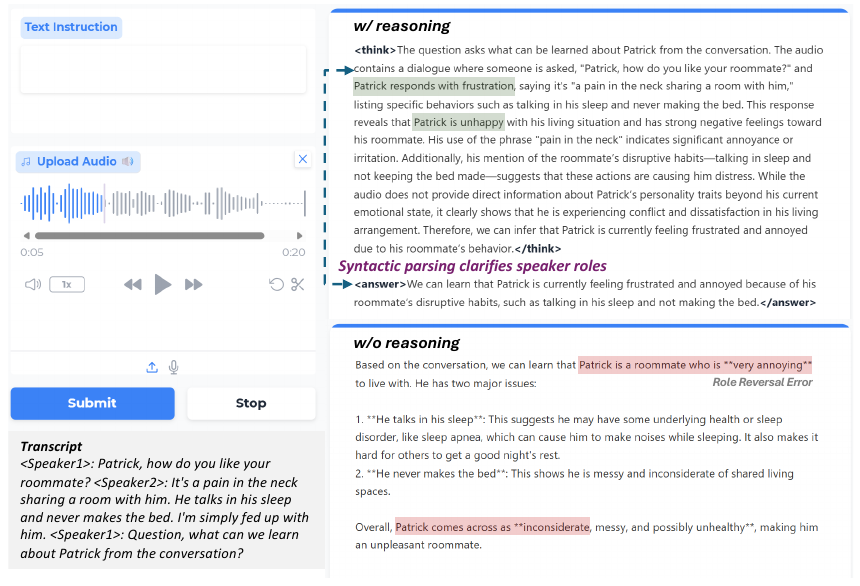}
    \caption{}
  \end{subfigure}
  \caption{\textbf{Success and failure modes in multimodal reasoning.} (a) The M2O-ML-r1 model hallucinates invisible physiological processes in the chest X-ray to satisfy its logical narrative. (b) The M2O-ML-r1 model successfully disentangles speaker intent in a complex audio dialogue by explicitly tracking speaker states.}
  \label{fig:6}
\end{figure}

\subsubsection{Qualitative Diagnostics: Hallucination \& Logic}
In this subsection, we further investigate the mechanisms behind both success and failure through a token-level analysis in Figure~\ref{fig:4} and qualitative case studies in Figure~\ref{fig:6}.

\begin{itemize}
    \item \textbf{The Mechanism of Drift:} In Figure~\ref{fig:4}, we trace the source of M2O-ML-r1's performance degradation on SEA-Video Benchmark to a structural conflict. Unexpectedly, we observe a striking anomaly during Audio-QA where M2O-ML-r1 generates up to a 6-fold surge in visual anchor tokens compared to the M2O-ML model (see the brown and red bars in Figure~\ref{fig:4}b). We term this phenomenon ``Visual Distraction'' during multimodal reasoning. What stands out is that even when we explicitly prompt for audio, the model’s Chain-of-Thought still ``hallucinates'' visual relevance.
    We hypothesize that this ``Visual Over-processing'' fills the limited context window with excessive spatial information. This interferes with temporal acoustic signals and may lead to what we call ``Temporal Drift''.
    \item \textbf{The Visual Failure:} As shown in Figure~\ref{fig:6}a, the X-Ray case demonstrates ``\textbf{Logical Over-fitting}'' intuitively. The model correctly identifies lungs but hallucinates reduced oxygen exchange, which is invisible in the image, to satisfy the model's medical narrative.
    \item \textbf{The Audio Success:} Conversely, in the ``Patrick the Roommate'' case in Figure~\ref{fig:6}b, our M2O-ML-r1 model succeeds where the baseline fails. By reasoning about speaker roles (``\textit{Patrick responds with frustration...}''), the model resolves syntactic ambiguity that puzzles the perception-only baseline.
\end{itemize}

\section{Related Work}
\label{sec:related}
The development of foundation models has gradually shifted away from English to multilingual systems designed for specific linguistic regions, including Southeast Asia~\citep{nguyen2024seallms, tjhi-etal-2023-sea, Maria2024Compass}.
A number of recent works have introduced LLMs for low-resource languages in Southeast Asian, including Thai, Vietnamese, and Indonesian. Most of these improvements come from curated regional corpora, more suitable tokenization schemes, as well as instruction tuning aligned with local cultural patterns~\citep{petrov2023language, Zhang2025SeaLLMs3}. Nevertheless, most previous research remains linguistic. The global research focus is aggressively shifting toward omni-modal integration of video, audio, and text~\citep{han2024onellm, xu2025qwen2, xu2025qwen3}. However, comparable multimodal integration is still limited in regional models, which are often developed around isolated modalities.
For instance, current Southeast Asian models tend to specialize in isolated modalities, with most work concentrating on either speech recognition~\citep{wang2025advancing, he2025meralion} or static image captioning~\citep{cahyawijaya-etal-2025-crowdsource}.
This specialization makes it harder for these systems to model joint visual–acoustic dynamics of visual and acoustic signals that characterize everyday communication in the region. Because of this, phenomena like code-switching or culturally specific visual cues remain poorly represented. These gaps are not only absent from text-centric regional models, but are also inadequately handled by Western-developed multimodal systems that are trained without sufficient regional grounding~\citep{pouget2024no, kadiyala2025uncovering, liu2025culturevlm}.

\section{Conclusion}
\label{sec:conclusion}
In this report, we introduced \textbf{MERaLiON2-Omni} (Alpha) and explored how System 2–style reasoning can be transferred to multimodal models using a resource-efficient silver-data pipeline. Our experiments show that reasoning improves performance on abstract tasks, including text-based logic (+11\%) and over 10\% improvement in spoken instruction following, but the same mechanism can destabilize low-level perceptual processing. Our analysis shows that reasoning aids visual understanding but can fail in long audio contexts, causing temporal errors and over-inference. Since scaling reasoning alone won't make a robust omni-model, our future work will explore adaptive routing between fast perception and slow deliberation.

\section{MERaLiON Team}
Zhang Longyin, Sun Shuo, He Yingxu, Won Cheng Yi Lewis, Muhammad Huzaifah Bin Md Shahrin,  Sailor Hardik Bhupendra, Wong Heng Meng Jeremy, Tarun Kumar Vangani, Ma Yi, Wang Qiongqiong, Pham Minh Duc,  Jiang Ridong, Li Jingtao, Liao Jingyi, Liu Zhuohan, Lu Yanfeng, Manas Gupta, Aw AiTi

\bibliography{biblio}
\bibliographystyle{colm2024_conference}

\end{document}